\documentclass[11pt,a4paper]{article}
\usepackage[hyperref]{acl2021}
\usepackage{times}
\usepackage{latexsym}
\usepackage{amsmath}
\usepackage{graphicx}

\usepackage{microtype}

\aclfinalcopy 

\title{Edeflip: Supervised Word Translation between English and Yoruba}

\author{Ikeoluwa Abioye \\
  \texttt{Ike.23@dartmouth.edu} \\\And
  Jiani Ge \\
  \texttt{Jiani.Ge.23@dartmouth.edu} \\}

\date{}

\begin{document}
\maketitle
\begin{abstract}
In recent years, embedding alignment has become the state-of-the-art machine translation approach, as it can yield high-quality translation without training on parallel corpora. However, existing research and application of embedding alignment mostly focus on high-resource languages with high-quality monolingual embeddings. It is unclear if and how low-resource languages may be similarly benefited. In this study, we implement an established supervised embedding alignment method for word translation from English to Yoruba, the latter a low-resource language. We found that higher embedding quality and normalizing embeddings increase word translation precision, with, additionally, an interaction effect between the two. Our results demonstrate the limitations of the state-of-the-art supervised embedding alignment when it comes to low-resource languages, for which there are additional factors that need to be taken into consideration, such as the importance of curating high-quality monolingual embeddings. We hope our work will be a starting point for further machine translation research that takes into account the challenges that low-resource languages face.
\end{abstract}

\section{Introduction}
In recent years, alignment of monolingual embedding spaces has replaced training on parallel corpora as the state-of-the-art machine translation approach \cite{word2vectrans, fbmuse}. Embeddings are representations of language tokens (words, sub-words, phrases) in a high-dimensional vector space. They represent the semantic and syntactic relationships between the tokens by learning the contexts in which they appear. As similar semantic and syntactic relationships exist in many languages, embedding alignment exploits this regularity to learn a transformation matrix that matches the relative positions of tokens with similar meanings in their respective monolingual embedding spaces \cite{word2vectrans, fbmuse}. As of today, continued refinement of the alignment algorithm has led to translation performance that either matches or exceeds that of the previous models based on parallel corpora \cite{fbmuse, artetxe-etal-2019}. Embedding alignment has the advantage of reducing the need for multilingual parallel corpora, which are costly to curate and limited in availability for many language pairs. However, most of the relevant research and published resources to-date focus on high-resource languages, such as English, Spanish, or Chinese. Even though the reduced need for parallel corpora promises increased accessibility for the translation of low-resource languages, it is unclear how they would actually fare under this approach.
 
The present study addresses this question by focusing on word translation between English and Yoruba. Yoruba is a language of the Niger-Congo family with over 50 million speakers in West Africa, though under-resourced in NLP research \cite{joshi-etal-2020}. As \citet{alabi2020massive} points out, the available monolingual resources for Yoruba, such as the pre-trained fastText embeddings, are of very low quality. First, many tokens in these embeddings are missing diacritics or marked incorrectly. In Yoruba, diacritics mark distinct letters and tones and are crucial to meaning representation. Second, these embeddings contain many English or other non-Yoruba tokens, which may bias the vector space. Finally, they are many times smaller in size (the pre-trained Wikipedia fastText Yoruba embedding contains 21,730 tokens, whereas the English one contains more than 2 million \citep{mikolov2018advances}).

As embedding alignment relies on monolingual embeddings, it is unclear how small and noisy embeddings, which are often the case for low-resource languages, affect translation performance. In this study, we adopt the supervised embedding alignment method from \cite{fbmuse}, manipulate the quality of the Yoruba embedding, and examine its effect on word translation precision. Investigating how the available embeddings for low-resource languages perform in embedding alignment may help us assess the extent to which the current state-of-the-art machine translation approach immediately benefits these languages, as well as the factors that need to be addressed to obtain high-quality machine translation results.

\section{Methodology}
\subsection{Data}
We use data for three purposes in the study. All data are collected from online, published sources.
\paragraph{Monolingual embeddings:} We use existing English and Yoruba embeddings for the alignment. We use a pretrained fastText embedding for English, with 1 million word vectors trained on Wikipedia and web crawl data \cite{mikolov2017advances}. For Yoruba, we use a pretrained fastText embedding, with 21,730 word vectors trained on Wikipedia \citep{bojanowski2017enriching}, and a second embedding trained by \cite{alabi2020massive}, with 113,572 word vectors trained on high-quality texts from Yoruba news and websites with complete diacritics notations. We use word embeddings, rather than embeddings with subword tokens, because the latter is not available for the high-quality Yoruba corpora.
\paragraph{Ground-truth dictionary:} For the ground-truth dictionary used in supervised alignment, we scrape the online English-Yoruba dictionary \citet{alabi2020massive} and produce 3,693 unique word pairs. We create separate entries for words with more than one translation. Entries with multi-word translations are excluded, as the embeddings we use only contain one-word tokens.
\paragraph{Evaluation dictionary:} For evaluating translation precision, we take the word pairs from the WordSim353 dataset translated into Yoruba by \cite{alabi2020massive}. We remove entries with multi-word translations and out-of-vocabulary words, and obtain 161 unique word pairs.

\subsection{Algorithm}

We adapt our algorithm from Facebook MUSE research project \cite{fbmuse} on supervised machine translation. \cite{fbmuse} most recently developed an unsupervised alignment method that uses adversarial learning to create a synthetic parallel dictionary, then applying supervised Procrustes alignment to the embeddings, using the synthetic dictionary as anchor points. Procrustes alignment matches the relative positions of two embedding spaces by finding the optimal orthogonal matrix that rotates, scales, and transforms them to be mostly closely aligned. The Procrustes problem has a closed-form solution through Singular Value Decomposition (SVD): \[ W^* = \underset{W\in{O_d(R)}}{\arg\max}{||WX - Y||_F = UV^T} \]
Where $U\Sigma V^T = \text{SVD}(YX^T )$, $W$ is the mapping or transformation matrix, $X$ is the source language embedding space, and $Y$ is the target language embedding space.

Due to the high computational demand of the adversarial learning phase of the method, we decide to perform Procrustes alignment using a human-created ground-truth dictionary as anchor points, thus rendering the overall process supervised. In half of the trials, prior to the alignment, we center and normalize the embeddings using the tensor operations from the PyTorch package, in order to make the distributions of the two embedding spaces more similar.

\subsection{Experiment}

We create four pairs of aligned English and Yoruba embeddings by manipulating two variables. Two of the alignments normalize the embeddings before alignment while the other two do not; and two of the alignments use the Wikipedia Yoruba embedding, while the other two the curated Yoruba embedding. Hereafter, the four conditions will be referred to as \textbf{wiki-unnormalized, wiki-normalized, curated-unnormalized,} and \textbf{curated-normalized}. We evaluate the alignment result by calculating the English to Yoruba word translation precision based on the evaluation dataset. As word translation is generated by retrieving the nearest neighbors of the source word in the target language embedding space, we calculate the probabilities that the top 1, 5, and 10 retrievals (k=1, 5, and 10) contain the correct translation. For example, a precision of 0.3 at k=5 means that for 30 percent of source words, the correct translation (or one of the correct translations) appears in the top five nearest neighbors in the target language embedding space.

\section{Results}
Table \ref{resultstable} shows the word translation precision of the four alignment conditions. Overall, the wiki-unnormalized condition consistently yields the lowest precision, whereas the curated-normalized condition the highest. Overall, the normalized alignments yield a higher precision than the unnormalized alignments that use the same embeddings. Rather than a main effect of embedding quality, however, our results show an interaction between embedding quality and normalization: whereas embedding normalization greatly improved the word translation precision for the curated conditions, the improvement for the wiki conditions is relatively small.

We also assessed the word translation results qualitatively by looking at individual queries and visualizing the aligned embeddings using t-SNE plots (Figures 1-4). The outcomes are consistent with the above-reported normalization and interaction effects. For example, given the source word ‘sea’, only the curated-normalized alignment yields the correct Yoruba translation 'òkun' (at k=3). The top retrievals of the wiki alignments are similar to each other and include omí ‘water’, ìrìnàjò ‘travel’, which are arguably related to ‘sea’. The top retrievals of the curated-unnormalized model, however, bares no visible semblance to the meaning of ‘sea’; they include, tápárebì ‘old age’ and káp\d{\`{o}}n\d{\'{o}}nlé ‘the only one.’ As can be seen in Figures 1-4, before normalization, the English and Yoruba embeddings are on different scales, with the Yoruba vectors being much farther apart. After normalization, the vectors of the two languages are on a similar scale, though they still do not constitute a perfect mapping.

\begin{table}[htbp]
\begin{center}
\begin{tabular}{||c c c c||} 
 \hline
 Condition & k = 1 & k = 5 & k = 10 \\ [0.5ex] 
 \hline\hline
 Wiki-unnorm & 10.56 & 18.01 & 21.74 \\ 
 \hline
 Wiki-norm & 12.42 & 19.25 & 25.47 \\
 \hline
 Curated-unnorm & 6.88 & 11.88 & 14.38 \\
 \hline
 Curated-norm & 19.38 & 23.75 & 29.38 \\[1ex]
 \hline
\end{tabular}
\end{center}
\caption{en-yo word translation precision at k=1, 5, 10}
\label{resultstable}
\end{table}

\section{Discussion}
\subsection{Word translation performance with en-yo}
Overall, the results are as expected. A comparison of the four alignment conditions shows clear effects of the quality of the embeddings and normalization. Using larger and correctly marked Yoruba embeddings and normalizing the embeddings before alignment increases the quality of the alignment and hence word translation accuracy. This is consistent with that the precisions we obtain are significantly lower than those in the published articles. For example, in \citet{fbmuse}, the same supervised alignment method produces a word translation accuracy at k=1 of 77.4 for English-Spanish, 68.4 for English-German, and 40.6 for English-Chinese. For one thing, these are high-resource languages and the embeddings that these alignments use are multitudes larger than the Yoruba embeddings we use. Further, the low precision that we obtain for English-Yoruba may in part extend from the trend that precisions are lower for language pairs that are more morpho-syntactically different, which are also suggested by the just-mentioned results for Spanish, German, and Chinese.

\begin{figure}[h!]
  \includegraphics[scale=0.33]{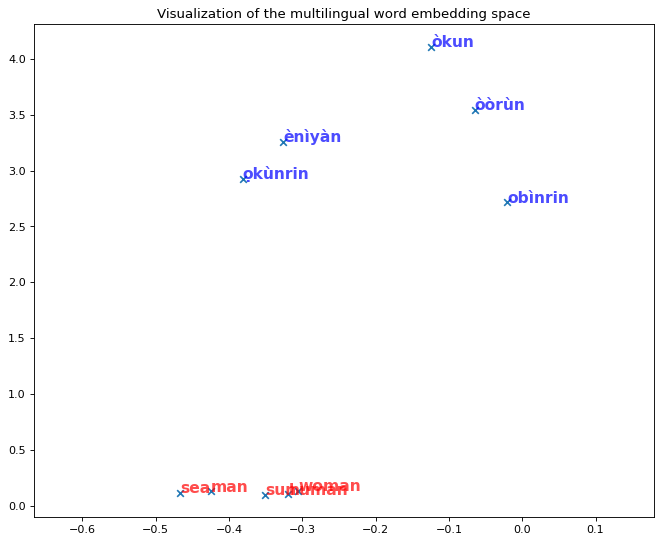}
  \caption{t-SNE plot for wiki-unnormalized}
  \label{fig:wiki_unnorm}
\end{figure}
\begin{figure}[h!]
  \includegraphics[scale=0.33]{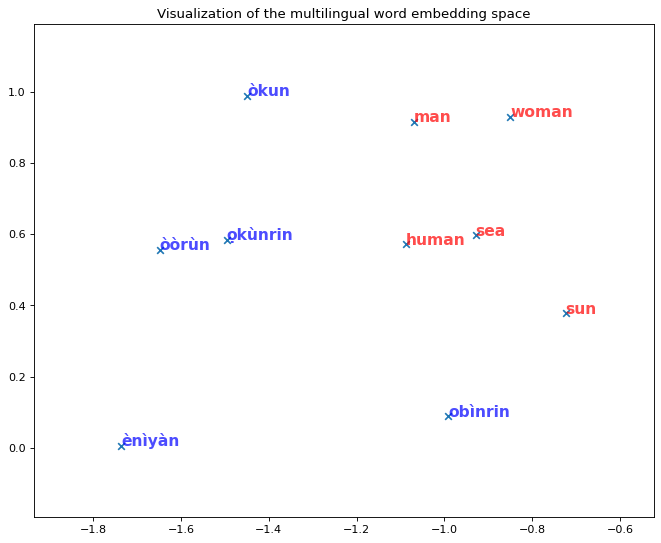}
  \caption{t-SNE plot for wiki-normalized}
  \label{fig:wiki-norm}
\end{figure}
\begin{figure}[h!]
  \includegraphics[scale=0.33]{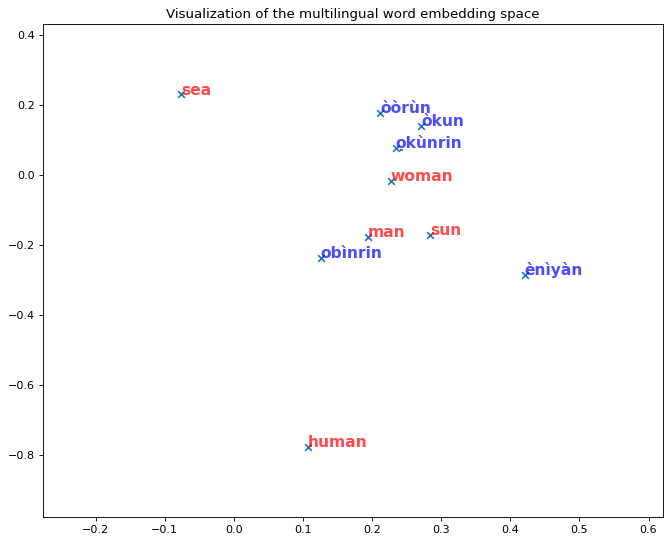}
  \caption{t-SNE plot for curated-unnormalized}
  \label{fig:curated_unnorm}
\end{figure}
\begin{figure}[h!]
  \includegraphics[scale=0.33]{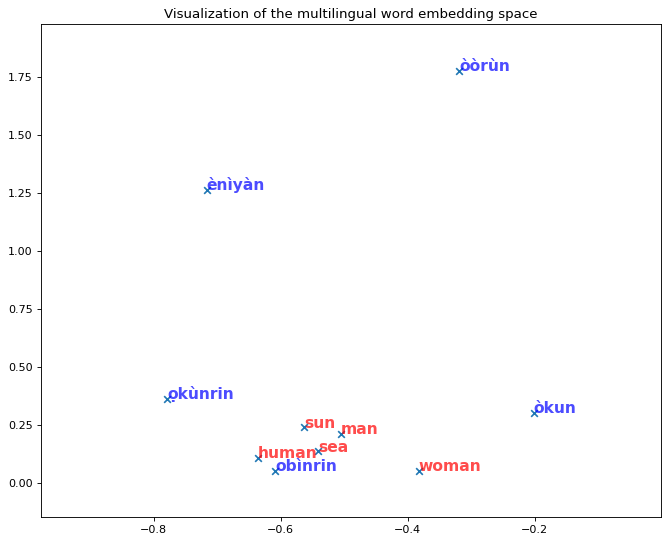}
  \caption{t-SNE plot for curated-normalized}
  \label{fig:curated_norm}
\end{figure}

The interaction effect between embedding quality and normalization is not predicted, though in hindsight understandable. We reason that it is because the English and Yoruba embeddings trained on Wikipedia have a more similar shape from the start, as they are both trained using fastText library, and their training data may have similar words and word frequencies. In comparison, the curated Yoruba embedding is trained independently on more distinct corpora \cite{alabi2020massive}. A similar phenomenon is reported in \citet{fbmuse}, where using the same alignment method, the word translation accuracy between English and Italian is higher when Wikipedia embeddings are used for both languages, compared to using Wikipedia embeddings for one language and  CBOW embeddings for another (p. 8). Nevertheless, in the curated-normalized condition in our study, the possible “disadvantage” of dissimilarity between source and target embeddings is offset by the normalization procedure and the superior quality of the curated Yoruba embeddings.
Taken together, our study shows that for word translation through embedding alignment, the quality (size and correctness) of the embeddings, normalization procedure, and the similarity between the original source and target embeddings all contribute to translation precision. While these factors may be less salient for translations between languages that have comparable high-quality embeddings, they raise crucial challenges to be addressed for projects that involve low-resource languages.

\subsection{Ethical considerations}
Our study also brings to attention several ethical considerations, primarily due to Yoruba's status as a low-resource language. Limited Yoruba NLP resource can inadvertently lead to exclusion and demographic bias. Compromised machine translation programs performance may risk poor translations being circulated without adequate verification. Such inaccuracies could result in miscommunication or misinterpretation, potentially leading to conflict and harm, especially considering the fact that while a translation may be technically accurate, it could also be culturally inappropriate or offensive. Furthermore, a significant disparity exists in the availability of computational resources; large organizations with substantial processing power can achieve superior results compared to smaller groups with limited resources. There is also an increased risk of privacy violations and exploitation of a community's linguistic resources without informed consent or tangible benefits.

\section{Conclusion}
Our study has shown that monolingual word embedding quality and normalization procedure affects the word translation performance of embedding alignment. Our findings identify challenges for machine translation with low-resource languages under the current state-of-the-art approach. Admittedly, the present study has many limitations. Future work may enhance the translation performance by using a larger training dictionary, incorporating more diverse tasks and metrics for evaluation, and, ideally, working with embeddings with even higher quality. We hope our work may be the starting point for a machine translation research that is more sensitive to the factors and challenges that particularly impact low-resource languages, and for efforts to address the resource inequality in the NLP.

\bibliographystyle{acl_natbib}
\bibliography{anthology,acl2021}


\end{document}